**Neuromorphic spatiotemporal optical flow: Enabling ultrafast visual perception beyond human capabilities**

**Authors**
Shengbo Wang[†1], Jingwen Zhao[†2], Tongming Pu[†1], Liangbing Zhao[†3], Xiaoyu Guo[†4], Yue Cheng[2], Cong Li[1], Weihao Ma[1], Chenyu Tang[5], Zhenyu Xu[6], Ningli Wang[7], Luigi Occhipinti[5], Arokia Nathan[8,9], Ravinder Dahiya[10], Huaqiang Wu[11], Li Tao[*2], Shuo Gao[*1]

**Affiliations**
[1]School of Instrumentation and Optoelectronic Engineering, Beihang University, Beijing 100191, China
[2]Center for Quantum Physics, Key Laboratory of Advanced Optoelectronic Quantum Architecture and Measurement (MOE), School of Physics, and Center for Interdisciplinary Science of Optical Quantum and NEMS Integration, Beijing Institute of Technology, Beijing 100081, China
[3]Computer Science, Computer, Electrical and Mathematical Science and Engineering Division, King Abdullah University of Science and Technology, Thuwal 23955-6900, Kingdom of Saudi Arabia
[4]Department of Mechanical Engineering, City University of Hong Kong, 83 Tat Chee Avenue, Hong Kong
[5]Department of Engineering, University of Cambridge, CB2 1PZ, Cambridge, UK
[6]Department of Precision Instrument, Tsinghua University, Beijing 100084, China
[7]Beijing Tongren Hospital, Capital Medical University, 100005, Beijing, China
[8]Darwin College, University of Cambridge, CB2 1PZ, Cambridge, UK
[9]School of Information Science and Engineering, Shandong University, Qingdao, 266237, China
[10]Bendable Electronics and Sustainable Technologies (BEST) Group, Department of Electrical and Computer Engineering, Northeastern University, Boston, MA, 02115, USA
[11]School of Integrated Circuits, Tsinghua University, Beijing 100084, China
[†]These authors contributed equally to this work
[*]Correspondence to: litao@bit.edu.cn, shuo_gao@buaa.edu.cn

**Abstract**
Optical flow, inspired by the mechanisms of biological visual systems, calculates spatial motion vectors within visual scenes that are necessary for enabling robotics to excel in complex and dynamic working environments. However, current optical flow algorithms, despite human-competitive task performance on benchmark datasets, remain constrained by unacceptable time delays (~0.6 seconds per inference, 4X human processing speed) in practical deployment. Here, we introduce a neuromorphic optical flow approach that addresses delay bottlenecks by encoding temporal information directly in a synaptic transistor array to assist spatial motion analysis. Compared to conventional spatial-only optical flow methods, our spatiotemporal neuromorphic optical flow offers the spatial-temporal consistency of motion information, rapidly identifying regions of interest in as little as 1–2 ms using the temporal motion cues derived from the embedded temporal information in the two-dimensional floating gate synaptic transistors. Thus, the visual input can be selectively filtered to achieve faster velocity calculations and various task execution. At the hardware level, due to the atomically sharp interfaces between distinct functional layers in two-dimensional van der Waals heterostructures, the synaptic transistor offers high-frequency response (~100 $\mu$s), robust non-volatility (>$10^4$ s), and excellent endurance (>8,000 cycles), enabling robust visual processing. In software benchmarks, our system outperforms state-of-the-art algorithms with a 400% speedup, frequently surpassing human-level performance while maintaining or enhancing accuracy by utilizing the temporal priors provided by the embedded temporal information.

## Main text

Optical flow, originally introduced by James J. Gibson in the 1950s and inspired by biological visual perception, estimates motion vectors within a visual scene (*1–3*). Over the decades of development, current state-of-the-art algorithms such as RAFT and GMFlow have demonstrated impressive performance on benchmark datasets. By leveraging two-dimensional gradient representations of pixel movement, optical flow offers clear, intermediate motion representations compared to end-to-end approaches, enabling it to excel in dynamic scene analysis at performance levels comparable to human capabilities (*4–9*). However, translating these achievements to real-world applications is still challenging due to the high computational overhead required to process visual inputs in real time. For instance, while Tesla's Autopilot employs occupancy networks to achieve latency as low as ~10 ms (*10*, *11*), performing optical flow analysis and object segmentation on a 1920 × 1080 resolution image can require over 0.6 seconds on an Nvidia V100 GPU, 4X humans (SNote 1 and 2). Such delays are unacceptable for time-sensitive applications like autonomous driving, where a one-second delay at highway speeds can reduce the safety margin by up to 27 meters, significantly increasing safety risks. Thus, due to this obvious time delay, the field deployment of optical flow seems to be unrealistic (*12–15*).

Optical flow is initially aimed to mimic the processing pipeline of biological visual systems but cannot faithfully replicate their high processing efficiency in practice, which is because biological vision excels at processing large volumes of visual information efficiently through dynamically focusing on regions where motion occurs (*16*, *17*). Critically, during this process, biological vision leverages not only spatial cues (e.g., gradient information captured by conventional optical flow) but also temporal cues, which analyse motion through temporal change information to achieve efficient visual processing. Replicating this dynamic processing in artificial systems is challenging due to the intrinsic inflexibility of conventional CMOS-based technologies, which cannot readily adjust their processing functions in response to varying stimuli (*18*, *19*). Fortunately, neuromorphic devices such as synaptic transistors and memristors offer synapse-like characteristics that can emulate the processing functions of biological synapses. As a result, these devices hold the potential to make human-like ultrafast visual processing possible (*20–23*).

Here, we propose a neuromorphic optical flow method that leverages two-dimensional neuromorphic synaptic transistors to form temporal motion cues directly in hardware, as shown in Fig. 1. By integrating these temporal motion cues with spatial gradient information, the resulting neuromorphic spatiotemporal optical flow offers a richer depiction of visual scenes than conventional spatial-only optical flow methods. As a result, the spatial-temporal consistency of the current visual scene can be captured by the synaptic transistors, enabling the rapid identification of potential motion regions within as little as 1–2 ms and ultimately accelerating various tasks such as motion prediction, segmentation, and tracking. Leveraging the superior properties of two-dimensional materials, such as atomic thickness and enhanced electrostatic control, the developed floating gate synaptic transistors demonstrate high-frequency response (~100 $\mu$s), robust retention (>$10^4$ s), and exceptional endurance (>8,000 cycles). In our experiments, we deployed the neuromorphic optical flow method across various application scenarios—including vehicle operation, UAVs, robotic arms, and sports activities—to perform tasks like motion prediction, object segmentation, and object tracking. The results demonstrate that our method significantly accelerates processing times, achieving an average ~400% speedup and surpassing human-level speeds (~150 ms) in most cases. Notably, by incorporating spatial-temporal consistency of motion information, our spatiotemporal approach maintains or improves accuracy, such as a 213.5% performance increase in the vehicle scenario. These advancements empower robots with ultrafast and accurate perceptual capabilities, enabling them to handle complex and dynamic tasks more efficiently than ever before.

**Floating gate synaptic transistor**

In neuromorphic optical flow, synapse arrays serve to embed temporal information from external visual scenes. To achieve precise encoding and long-term retention of this information, the synapse array must exhibit synapse-like characteristics—adjusting its state in response to external stimuli—and non-volatile properties to maintain stored data (*24*). To further ensure high-frequency processing capabilities and long-term system stability, we have designed floating gate synaptic transistors based on a two-dimensional van der Waals heterostructures as neuromorphic devices that generate temporal motion cues directly in hardware. Fig. 2a demonstrates a structure schematic of this floating gate synaptic transistor. From bottom to top, the synaptic transistor includes a gold film (serving as the control gate), an aluminium oxide ($Al_2O_3$) blocking layer, a multilayer graphene (MLG) floating gate, a thin hexagonal boron nitride (h-BN) tunnelling layer, and a molybdenum disulfide ($MoS_2$) channel. In operation, gate-source voltage ($V_{gs}$) pulses are applied to the control gate (with the source grounded) to modulate the drain-source current ($I_{ds}$). Comprehensive details on the fabrication process of the floating gate synaptic transistor and the Raman characterization of its heterostructure are provided in Figs. S1 and S2. Under these conditions, the $MoS_2$ channel's output characteristic confirms a good ohmic contact with the Cr/Au interface (Fig. S3). In terms of memory behaviour, the transfer curve of this synaptic transistor at a fixed drain-source voltage ($V_{ds}$=1 V) depicts a clockwise memory window that reaches 11.2 V when the $V_{gs}$ is swept from −10 V to +10 V (Fig. 2b). Furthermore, the memory window increases with the maximum applied $V_{gs}$, as presented in Fig. S4. When applying $V_{gs}$ pulses, this synaptic transistor displays obvious synapse-like characteristics. As shown in Fig. 2c, the change in conductance is positively related to the number of applied pulses. The modulation mechanism is elucidated by the energy band diagram: negative $V_{gs}$ pulses drive holes into the floating gate and elevate the device's conductance, while positive $V_{gs}$ pulses facilitate electrons tunnelling into the floating gate and reduce its conductance. Additional details about the operating mechanism can be found in Fig. S5. This modulation can be controlled by varying the amplitude and duration of applied voltage stimuli. As depicted in Fig. 2d, the increase in conductance correlates positively with both the amplitude and duration of the negative $V_{gs}$ pulses. Fig. S6 illustrates the variation of conductance with the amplitude of the positive $V_{gs}$ pulses. With respect to response speed, this floating gate synaptic transistor demonstrates rapid operating speed, achieving a current switching of 60 $\mu$A (from a low- to high-conductance state) under −15 V $V_{gs}$ pulse with a duration of 100 $\mu$s (Fig. 2e). This ~100 $\mu$s response time is suitable for high-frequency visual information processing. Moreover, the synaptic transistor exhibits repeatable programming characteristics and multiple analog states (Fig. 2f), enabling the precise encoding of external information as its state. In terms of endurance, up to 8,000 programming/erasing cycles can be achieved under positive and negative $V_{gs}$ pulses (±15V, 1 ms), with the $I_{ds}$ at $V_{ds}$=1 V remaining at ~$10^{-9}$ A and $10^{-5}$ A in the low- and high-conductance states, respectively (Fig. 2g). Furthermore, this synaptic transistor displays non-volatile behaviour. Both the low- and high-conductance states are maintained for over $10^4$ s (Fig. 2h), confirming their non-volatility in storing external stimuli data. Compared with other reported devices (Fig. 2i), this synaptic transistor requires a low $V_{gs}$ amplitude for weight modulation and presents excellent retention (*25*–*34*). This high performance in endurance and reliability can be attributed to the integration of a defect-free h-BN tunnelling layer (*35*), reasonable gate coupling ratio (*29*), and the establishment of atomically sharp and flat interfaces (*25*, *35*–*37*); a detailed comparison is provided in Fig. S7 and STable 1. When scaling a single synaptic transistor to a 4×4 array as shown in Fig. 2j, the developed fabrication process can be found in the Methods section and Fig. S8. After encapsulation (Fig. 2k), the array can be interfaced with external circuits via pins or connectors, facilitating integration with other system components. Such scalability paves the way for the development of commercial chips. The variation among multiple devices, as exhibited in Fig. 2l, demonstrates the consistent synaptic modulation behaviour.

**Temporal motion cue generated by neuromorphic devices**

To directly generate temporal motion cues at the hardware level, we propose the imaging architecture illustrated in Fig. 3a. In this design, a conventional imaging array could serve as the front-end sensor, converting external stimuli into an analog voltage signal. This signal is then processed along two parallel paths: one is digitized to form the conventional image representation, while the other converses the analog signal and modulates a synapse array to record temporal information. Specifically, the circuit needed for voltage conversion, shown in Fig. 3b and 3c, consists of a differential processing part and an amplitude conversion part. The differential processing part extracts changes in light intensity, while the amplitude conversion part generates voltage pulses applied to the synaptic transistor array, reflecting the temporal information of the current visual scene. In the differential processing part, a high-pass filter first differentiates the light intensity, and an operational amplifier (op-amp) then amplifies these changes within a suitable operating range for subsequent processing (Fig. 3b and Eq. 1):

$$V_{i,j}(t) = a\left|\frac{dI_{i,j}}{dt}\right| = a\left|I_{i,j}(t) - I_{i,j}(t-\Delta t)\right| \tag{1}$$

where $V$ is the output voltage of the differential processing unit; $\Delta t$ refers to the sampling interval; $(i, j)$ represents the spatial coordinates of the temporal information, corresponding in size to the synapse array; $I$ is the analog visual voltage transmitted from the imaging array (which may be resized to match the synapse array size); and $a$ is a proportionality coefficient ensuring the voltage remains within a suitable operating range. In the following amplitude conversion part, an absolute circuit is constructed to extract the absolute voltage change, focusing on the magnitude of the light intensity change rather than its direction (Eq. 2):

$$\hat{V}_{i,j} = \left|V_{i,j}\right| \tag{2}$$

Then, a reconfigurable op-amp generates a corresponding modulation pulse based on the current amplitude of the absolute voltage $\tilde{V}_{i,j}$. The relationship follows Eq. 3:

$$\tilde{V}_{i,j} = \begin{cases} plus_1 \times (\hat{V}_{i,j} - bia_1) & \hat{V}_{i,j} > V_{\mathrm{th}} \\ plus_2 \times (\hat{V}_{i,j} - bia_2) & \hat{V}_{i,j} \leq V_{\mathrm{th}} \end{cases} \tag{3}$$

where $\tilde{V}_{i,j}$ is the modulation voltage $V_{gs}$ applied to synapse array, $plus_1$, $plus_2$ are different proportional coefficients, $bia_1$, $bia_2$ represent different modulation biases, and $V_{th}$ is a preset threshold. Under the effect of modulation voltage $V_{gs}$, the synaptic transistor at position $(i, j)$ of array is modulated to a resistance state related to the nature of light intensity change. By comparing it with the threshold $V_{th}$, dramatic changes caused by potential moving objects and mild changes caused by background movement or noise are separated and translated into negative and positive $V_{gs}$ pulses, respectively, resulting in different trends of device state switching. When analyzing this synapse array, the temporal motion information can be inferred from the distribution of $I_{ds}$ values. For example, devices with high $I_{ds}$ values (low-resistance under the modulation of negative $V_{gs}$ pulses) that cluster spatially indicate regions containing moving objects. More details of this circuit and the manipulation of analog visual voltage can be found in Fig. S10.

In our implementation, we employed a commercial camera to capture 800×800 images within a vehicle scenario and processed the visual input using the 4×4 synapse array (Fig. 3d). During processing, the visual stimuli captured by the commercial image sensor are resized by averaging the light intensity of a matrix of $m \times n$ pixels into a basic unit. Here, $m$ and $n$ are set to 200. In this configuration, external visual stimuli are translated into modulation voltages for the synapse array according to the voltage conversion circuit. Thus, temporal information of current visual scene is mapped onto this transistor array. For instance, when a pedestrian suddenly runs in front of the car (Figs. 3d and 3e), a noticeable change in the analog visual voltage occurs at position (2, 4), leading to a negative $V_{gs}$ pulse. As a result, the synaptic transistor at (2, 4) switches to a high-conductance state, causing its $I_{ds}$ value to increase significantly under a fixed $V_{ds}$. As shown in Fig. 3f, the distribution of measured $I_{ds}$ across the 4×4 synaptic transistor array clearly reflects the temporal dynamics of the current visual scene, highlighting the presence of a moving object on the right side. To further facilitate integration with conventional visual processing methods, these temporal data are transformed into values ranging from 0 to 255 as shown in Fig. 3g using a logarithmic mapping (see Methods), making them compatible with common image processing libraries such as Python's OpenCV (cv2) package. Through this pipeline, the temporal information encoded by the neuromorphic devices can be seamlessly combined with the original image (Fig. 3h).

**Accelerated movement velocity calculation**

After extracting temporal information of the current visual scene from the synapse array states, this data can be transformed into temporal cues that accelerate velocity estimation, ultimately producing the neuromorphic optical flow (Fig. 4a). Specifically, the conversion into temporal cues involves two main steps. First, there is a binarization process based on a predefined threshold. Second, a connectivity analysis that includes defining connectivity, labeling connected components, and expanding these regions is performed (Fig. 4b). The resulting list of connected regions serves as the temporal cues, as shown in Fig. 4c. In addition to the vehicle operation scenario, the UAV operation scenario with a resolution of 160×160 is also processed using the same 4×4 synaptic transistor array. By comparing the original image and the generated temporal cues, it is observed that the constructed temporal cues areas effectively highlight potential moving objects. During the subsequent velocity inference, the temporal cues serve as regions of interest (ROIs) that help automatically filter the areas where movement velocities need to be calculated (Fig. 4d). This filtering process speeds up velocity calculations compared to processing the entire image. Additionally, neuromorphic optical flow seamlessly integrates with current velocity inference approaches, whether they are based on traditional computer vision techniques or neural networks.

In our implementation, we demonstrate the adaptability of our method using three representative algorithms for movement velocity calculation: the traditional Farneback algorithm and the neural network-based GMFlow and RAFT algorithms, as shown in Fig. 4e. Integration with other algorithms, such as FlowFormer, is demonstrated in Fig. S11 and S13. These algorithms vary in their operational characteristics and should be selected based on the practical working environment. For instance, Farneback is suitable for less demanding scenarios, while GMFlow and RAFT are more appropriate for situations requiring higher accuracy and adaptability, albeit at increased computational and technical costs. Nevertheless, the capability of integrating various movement velocity calculation methods ensures that the neuromorphic optical flow method is applicable to unstructured environments, which can vary significantly. Taking the vehicle operation scenario as an example, the temporal cues filter the visual input so that only highlighted regions are forwarded to the subsequent velocity inference stage. In practical implementations, to improve robustness and resolution, the selected region is slightly padded to include peripheral information about the moving area. As shown in Fig. 4f, the detected motion region is slightly padded, and the movement information can be calculated using various velocity inference algorithms. Details on the strategy of padding and handling multiple moving objects can be found in SNote 5.

In the vehicle scenario, processing examples are shown in Fig. 4g. Even during periods of high motion—when the running pedestrian occupies a large portion of the scene—our method remains faster (Fig. 4i). Compared to conventional optical flow methods, our spatiotemporal approach enables the detection of potential motion regions (1–2 ms) as shown in Fig. S15, and the average total velocity inference times for Farneback, GMFlow, and RAFT are reduced to 13.0%, 37.2%, and 19.6%, respectively (Fig. 4j). For the UAV scenario analysed with Farneback as shown in Fig. 4h, the average inference time is reduced to 51.0% of the original duration (Fig. S12).

**Fundamental tasks empowered by neuromorphic optical flow**

Neuromorphic optical flow, which integrates temporal and spatial motion cues (e.g., calculated spatial movement velocity), supports fundamental tasks that enable autonomous vehicles, UAVs, and robots to perceive, understand, and interact with their environments autonomously and intelligently (Fig. 5a). During task execution, these motion cues are first decomposed. Then, temporal motion cues selectively filter spatial cues to focus only on regions with potential motion information rather than processing the entire scene. This selective focus significantly accelerates subsequent task execution. The filtered spatial cues are then combined with visual input to execute task-specific algorithms. The implementation pipeline is detailed in SNote 4. Aligning with the vision of utilizing optical flow to help robotic systems efficiently perceive the dynamic world similar to biological visual systems, the above pipeline for processing visual information with neuromorphic optical flow mirrors the human visual system. This includes a perception unit corresponding to the human eye, a synapse array that extracts temporal motion cues analogous to the lateral geniculate nucleus (LGN), and task-specific algorithms that perform high-level processing similar to the visual cortex. Additional descriptions of the human visual system can be found in SNote 1.

In our implementation, the visual inputs encompass various scenarios, including vehicle operation, UAV operation, sports activities (e.g., table tennis from the UCF dataset) (*38*), and grasp operation. The settings for visual processing are provided in the Methods section and in STable 5. Utilizing neuromorphic optical flow, essential tasks such as motion prediction, object segmentation, and object tracking are performed on these visual inputs. Detailed task-specific algorithms can be found in the Methods section and Fig. S16. As shown in Fig. 5b, spatial motion cues are calculated using multiple velocity inference methods and employed to execute various tasks. Additional results are presented in Figs. S17–S21. The overall accelerated processing, which includes both velocity inference and task execution based on neuromorphic optical flow, achieves processing times comparable to human perception—approximately 150 ms (SNote 1). In addition to high processing efficiency, these tasks are evaluated using standard metrics, including the Structural Similarity Index Measure (SSIM), pixel accuracy (PA), and Intersection over Union (IoU), and neuromorphic optical flow achieves comparable performance to conventional optical flow (Fig.

6a). In certain scenarios, such as vehicle operation, UAV operation (small), and grasping, neuromorphic optical flow significantly outperforms conventional methods. On average, the accuracy improvements are 213.5%, 157.4% and 740.9%, respectively. These performance improvements are attributed to the additional environmental knowledge embedded in the temporal cue. For instance, in the RAFT-based object segmentation for grasping operations (Fig. 5b), RAFT cannot infer velocity accurately due to its limited generalization. However, the temporal cue provides a boundary constraint, enhancing segmentation accuracy. Similarly, in object tracking tasks, the temporal cue highlights the region containing the moving object while excluding irrelevant regions. As a result, the tracking accuracy is improved by reducing the impact of noise. More detailed task performance data can be found in STable 2.

In terms of processing efficiency (Fig. 6b), the ability of the neuromorphic optical flow to accelerate full-spectrum visual processing, including both velocity inference and task-specific algorithms. Thus, a faster response can be observed compared to conventional optical flow methods. When using Farneback for velocity inference, the acceleration ratio—i.e., the ratio of the original whole processing time to the whole accelerated processing time—ranges from 12.5% to 58.0%, with an average of 27.5%. For GMFlow-based tasks, the acceleration ratio ranges from 4.7% to 36.7%, with an average of 20.6%. For RAFT-based tasks, the acceleration ratio ranges from 16.7% to 53.3%, with an average of 29.1% (STable 3). After examining all 33 groups of tasks, the average acceleration ratio is 26.1%, which corresponds to an approximate 4X speedup. When fitting the acceleration ratio of velocity inference and the percentage of filtered regions based on temporal cues in the vehicle scenario, clear linear relationships are observed for both Farneback and RAFT methods, with all $R^2$ values exceeding 0.94 (Fig. 6c). However, velocity inference using GMFlow does not follow this trend due to its unique operational characteristics (Fig. S14). Other velocity inference methods, such as FlowFormer, exhibit similar linear acceleration trends (Fig. S13). In the acceleration of task execution, similar linear relationships can be observed using data points from Farneback-based vehicle scenario processing. This general acceleration, which correlates with the size of filtered visual input based on temporal cues, demonstrates the effectiveness of the proposed approach in enhancing both velocity inference and task execution performance. As a result, neuromorphic optical flow enables real-time visual processing capabilities that are comparable to, or even exceed, human-level perception (Fig. 6d and STable 4).

## Discussion

### *Comparison with conventional optical flow methods*

Compared to conventional spatial-only optical flow methods, neuromorphic spatiotemporal optical flow integrates additional temporal motion cues of the current visual scene. By utilizing the spatial-temporal consistency of motion, which refers to the simultaneous spatial displacement of pixels and the temporal variation in light intensity within a motion region, this added temporal information enables the direct delineation of potential moving regions in as little as 1–2 ms using synapse array state information. This delineation offers two major benefits. First, it enables selective processing of visual input, resulting in substantially faster velocity calculations and task execution. Second, the delineation information provides valuable prior knowledge for velocity inference and task execution processes. For instance, in object tracking tasks, the temporal cues from neuromorphic optical flow constrain the tracking range, reducing false detections from background noise and greatly enhancing robustness. Furthermore, for neural network–based velocity inference, the delineation information supplies a reasonable range of results even in untrained working environments, thus addressing the limited generalization problem of current neural network methods.

***Benefits in practical implementation***

Neuromorphic optical flow significantly reduces the processing time of visual data, enabling robotics to excel in more complex tasks, particularly those requiring real-time processing capabilities like collision avoidance and object tracking. For example, in vehicle operations, the average ~0.2 seconds improvement in processing time observed in our method can result in a reduction in full-braking distance of 4.4 meters at a speed of 80 km/h, greatly enhancing driving safety. Similarly, our method enables at least a threefold reduction in reaction time in UAV(small) scenarios, significantly improving their durability and performance in dynamic environments. Across all tasks using Farneback and GMFlow for velocity inference, processing times remain below 40 ms. This enhanced processing allows UAVs to track moving objects between frames in the setting of 25 frames per second. As a result, UAVs can adjust their speed and pose in real-time, achieving near-theoretical minimum delay in target tracking. Beyond robotic applications, neuromorphic optical flow holds great promise for improving human-robot interaction. With an emphasis on response time to ensure real-time feedback, robots must interpret visual scenes—such as gestures and movement recognition—within 100 to 200 ms. The ultrafast visual processing enabled by neuromorphic optical flow can serve as a crucial information source for future human-robot interaction, ensuring smooth and responsive engagements.

***Future directions***

The core principle of neuromorphic optical flow lies in capturing the temporal information of visual scenes through synapse arrays, enabling a temporally guided analysis of visual stimuli. This design ensures high compatibility with various types of front-end sensors. Beyond the currently demonstrated image-based sensors, event-based cameras present a promising avenue for integration. Event-based cameras are uniquely suited to address challenges faced by image-based methods, such as motion blur and over-saturated image regions caused by the limited dynamic range of traditional sensors. By incorporating event-based cameras as the front-end sensor, the synapse array in the neuromorphic optical flow can be modulated by event streams, potentially delivering similar acceleration benefits for event-based optical flow methods. This integration could endow robotic systems with significantly enhanced perceptual capabilities. In terms of applications, this temporally guided approach extends far beyond optical flow calculation alone. For instance, after identifying potential ROI within our proposed system architecture, other algorithms—such as YOLO neural networks for object detection—can be directly applied to these identified areas. As a result, computational resource usage and the time required for visual motion processing are minimized. With the ability to enhance efficiency across a wide range of applications, our spatial-temporal integrated approach could pave the way for broader adoption in fields such as robotics, autonomous systems, and computer vision, driving transformative advancements.

In conclusion, this work proposes and demonstrates a neuromorphic spatiotemporal optical flow method leveraging synapse array to deliver a more comprehensive and efficient understanding of visual scenes compared to conventional spatial-only optical flow approaches. Compared to conventional optical flow methods, our method encodes additional temporal motion cues directly within the hardware, which identify regions of interest in real time. Therefore, the full spectrum of optical flow-based visual processing can be accelerated. Furthermore, the seamless integration of various movement velocity calculation algorithms ensures adaptability across real-world complex environments. Benchmark evaluations across multiple robotic platforms and tasks demonstrate that our neuromorphic optical flow outperforms state-of-the-art algorithms, achieving an average 4X improvement in processing speed while maintaining or enhancing accuracy in motion prediction, object tracking, and segmentation. Notably, our method facilitates the entire processing time, including velocity inference and task execution that approach or exceed human-level speeds (approximately 150 ms), thereby realizing the initial vision of optical flow and providing autonomous systems with unparalleled perception capabilities essential for safe and intelligent interaction with dynamic environments.

## Methods

**Device fabrication:** The bottom control gates in the floating gate synaptic transistors were prepared via electron beam lithography (EBL) on 285 nm $SiO_2$/Si substrate, followed by thermal evaporation. Atomic layer deposition (ALD) technique was used to deposit the gate dielectric aluminum oxide ($Al_2O_3$). $MoS_2$/h-BN/MLG heterostructure in single device was prepared by mechanical exfoliation from bulk materials (Nanjing MKNANO Tech. Co., Ltd.), and precisely positioned on the $Al_2O_3$ via dry-transfer method. $MoS_2$/h-BN/MLG structures in floating gate synaptic transistor array were fabricated from chemical vapor deposited materials (Six Carbon Technology Shenzhen), and stacked via wet transfer, then patterned through reaction ion etching according to the structural design. Furthermore, 5/50 nm Cr/Au drain-source electrodes on $MoS_2$ channel were defined sequentially by EBL, thermal evaporation and lift-off.

**Device characterization:** Electrical measurements of the floating gate synaptic transistor were conducted by a semiconductor parameter analyzer (Bl500A, Keysight) under atmospheric conditions. The thickness of the two-dimensional material was measured by Bruker Mulyi-Mode 8 AFM. Raman characterization was tested using a confocal Raman spectrometer (WITec alpha 300R) with a 532 nm laser as the excitation source.

**Logarithmic mapping:** To transform the drain-source current of the floating gate transistor to a range of 0–255, a logarithmic mapping is employed as shown in Eq. 4:

$$s=-\frac{3366}{\log_{10}^{I_{ds}}}-306 \tag{4}$$

where $I_{ds}$ represents the drain-source current, and $s$ is the transformed current scaled within the range of 0 to 255.

**Algorithms designed for motion prediction, object segmentation, and object tracking:** In the task of motion prediction, it involves predicting the position of a moving object. For the filtered visual input, by using reference frames and employing the Lanczos interpolation method, the moving object in the next moment can be inferred. The Lanczos interpolation formula is as Eq. 5 and 6:

$$g(x)=\sum_{k=-n}^{n} g(k)L_n(x-k) \tag{5}$$

$$L_n(x)=\begin{cases}\dfrac{\sin(\pi x)\sin(\pi x/n)}{(\pi x)^2}, & \textbf{\textit{if}}\ x\neq 0\\ 1, & \textbf{\textit{if}}\ x=0\end{cases} \tag{6}$$

where $g(x)$ represents the interpolated value at position $x$, $g(k)$ represents the pixel values at integer positions $k$ in the current frame, $L_n(x)$ is the Lanczos kernel function, and $n$ is the size of the kernel.

For object segmentation, the neuromorphic optical flow is first converted to polar coordinates. This transformation separates the original movement velocity information into direction (angle) and magnitude (distance) components, making it easier to analyze motion information. Notably, this step only manipulates the ROI that includes significant moving objects inferred from the motion pattern layer, thereby omitting regions with slight noise caused by environmental changes, such as slow variations in lighting. After the transformation, the image is converted from RGB to HSV color space, where the direction and magnitude of motion velocity layers are represented using the hue and value channels, respectively. This process is beneficial for subsequent processing because it allows more intuitive manipulation of color-based information: the hue channel encodes the direction of motion, while the value channel encodes the magnitude of motion. This separation simplifies the process of identifying and segmenting moving objects based on their motion characteristics. When the motion information of the ROI is represented in the HSV color space, thresholding operations along with erosion and dilation operations can be applied to create a binary mask that accurately segments the moving objects. Thresholding isolates the relevant motion information, while erosion and dilation help refine the segmentation by removing small noise and closing gaps in the detected objects, respectively. This process results in a clear and precise segmentation of moving objects, thus enabling subsequent tasks such as object tracking and interaction in dynamic environments.

In object tracking, coordinate conversion which is similar to that used in object segmentation is applied first. This conversion separates motion information into direction and magnitude components. Following this, morphological opening is performed to smooth boundaries and remove noise. Then, using the contour detection algorithm, multiple bounding boxes are detected. Next, non-maximum suppression is applied to eliminate redundant detections and retain the most significant objects. This step ensures that only the most prominent moving objects are tracked across frames, improving the accuracy of the tracking process. Unlike conventional optical flow, which can be disturbed by background movements leading to unnecessary tracking, neuromorphic optical flow focuses solely on the ROI regions that include potential moving objects. This targeted approach enhances tracking precision and reduces computational overhead by ignoring irrelevant background motion.

When evaluating the performance of the above tasks, metrics including structural similarity index measure (SSIM), pixel accuracy (PA), and intersection over union (IoU) are calculated to quantify the quality of prediction, segmentation, and tracking (Eq. 7-10), respectively:

$$\boldsymbol{SSIM}(x,y)=\frac{(2\mu_x\mu_y+C_1)(2\sigma_{xy}+C_2)}{(\mu_x^2+\mu_y^2+C_1)(\sigma_x^2+\sigma_y^2+C_2)} \tag{7}$$

where $\mu_x$ and $\mu_y$ are the average values of predict result $x$ and ground truth $y$, $\sigma_x^2$ and $\sigma_y^2$ are the variances of $x$ and $y$, $\sigma_{xy}$ is the covariance of $x$ and $y$, and $C_1$ and $C_2$ are constants to stabilize the division with weak denominator.

$$\boldsymbol{PA} = \frac{\sum_i n_{ii}}{\sum_i t_i} \tag{8}$$

where $n_{ii}$ is the number of pixels correctly classified for class $i$, and $t_i$ is the total number of pixels in class $i$. Here, PA calculates the percentage of correctly segmented pixels.

$$\boldsymbol{IoU} = \frac{\sum_{i=1}^{n} IoU_{si}}{n} \tag{9}$$

$$IoU_s = \frac{|A \cap B|}{|A \cup B|} \tag{10}$$

where $|A \cap B|$ is the area of overlap between the tracking mask $A$ and the ground truth mask $B$, and $|A \cup B|$ is the area of union between the tracking mask $A$ and the ground truth mask $B$, $IoU_{si}$ represents the $IoU_s$ between the $i$-th bounding box and the ground truth.

**Running environment:** Performance evaluations of neuromorphic optical flow with the Farneback method for velocity inference are performed on the 12th Generation Intel® Core™ i9-12900H processor. In contrast, performance evaluations of neuromorphic optical flow utilizing neural network–based velocity inference methods, including RAFT, GMFlow and FlowFormer, are conducted on a server outfitted with an NVIDIA V100 GPU and an Intel® Xeon® Platinum 8260 CPU operating at 2.40 GHz.

**Visual processing:** To demonstrate the scalability of neuromorphic optical flow, visual input data—encompassing UAV (small) operations, table tennis, and grasping scenarios—are simulated using the synapse array based on our fabricated synaptic transistor (Fig. S9).

**Visualization of optical flow:** Following the work by Baker et al., optical flow vectors are mapped to a color-coded image (*39*). In this visualization approach, color hue represents the direction of motion and color intensity/saturation corresponds to the magnitude of motion.

**Acknowledgements**

**Funding:** S.G. and L.T. acknowledges support from National Key Research and Development Program of China (2023YFB3208000, 2023YFB3208003 and 2023YFB3208002). L.T. acknowledges support from the Analysis & Testing Center at Beijing Institute of Technology, and the start-up fund provided by Beijing Institute of Technology.

**Author contributions:** S.W., J.Z., T.P., L.Z. and X.G. contributed equally to the work. S.G. and S.W. conceived the idea and proposed the research. S.W. and T.P. designed the neuromorphic optical flow pipeline, with T.P. developing task-specific algorithms. L.Z. and S.W. evaluated the neuromorphic optical flow across various scenarios. X.G. collected the datasets and conducted data pre-processing. L.T. conceived the floating gate synaptic transistor design. J.Z., Y.C. and L.T. fabricated, tested, and analyzed the synaptic transistors. S.G. and L.T. supervised the project. S.G., S.W., L.T. and J.Z. wrote the manuscript with inputs from all authors. All authors discussed the results and implications and commented on the manuscript at all stages.

**Competing interests:** The authors declare no competing financial interests.

**Data and materials availability:** The data that support the findings of this study are available from the corresponding authors upon request.

## Figures and Table

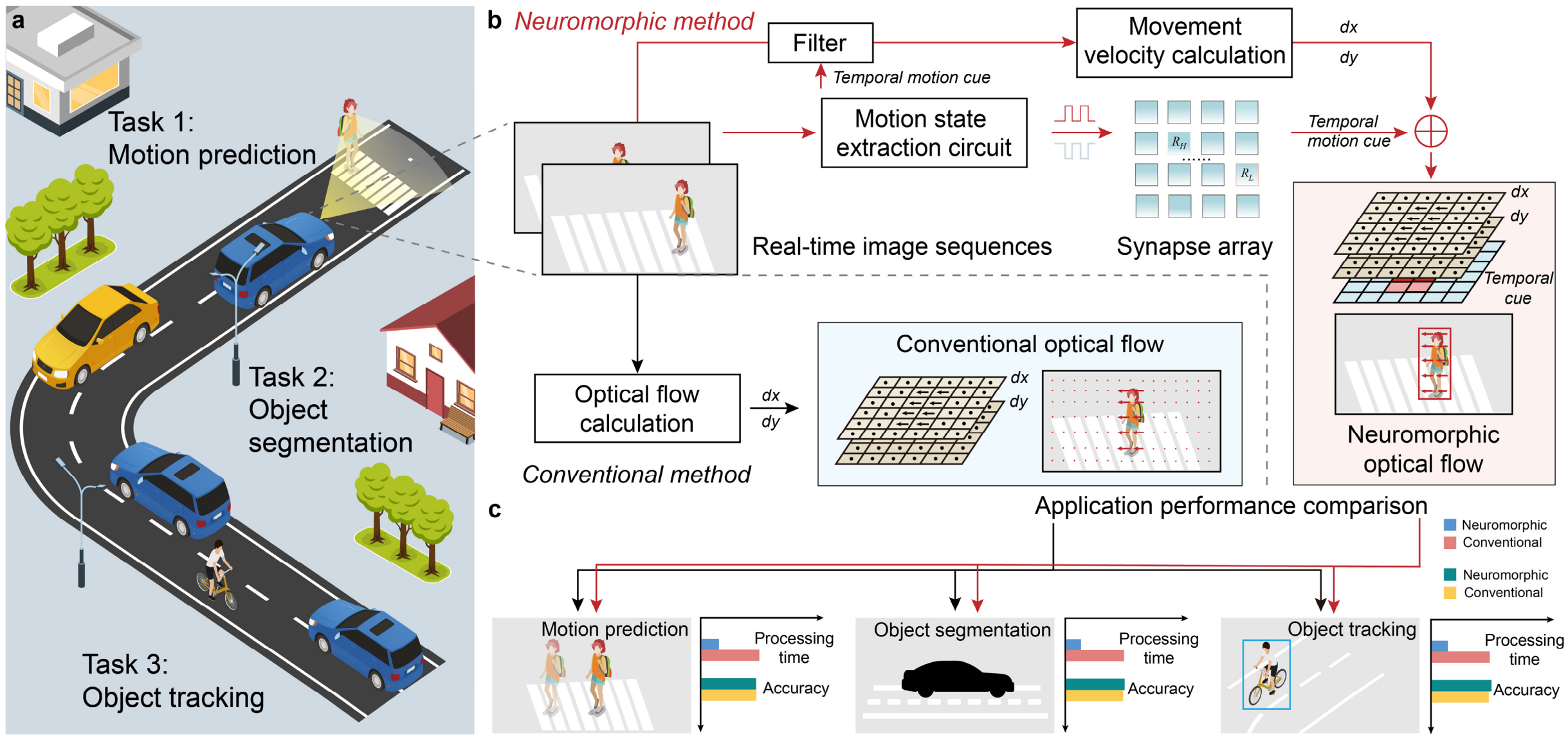

**Figure 1.** Schematic of the neuromorphic optical flow method and its application. **a**, Application scenarios for the neuromorphic optical flow method. **b**, Pipelines of the neuromorphic and conventional optical flow methods. **c**, Performance comparison between the two methods.

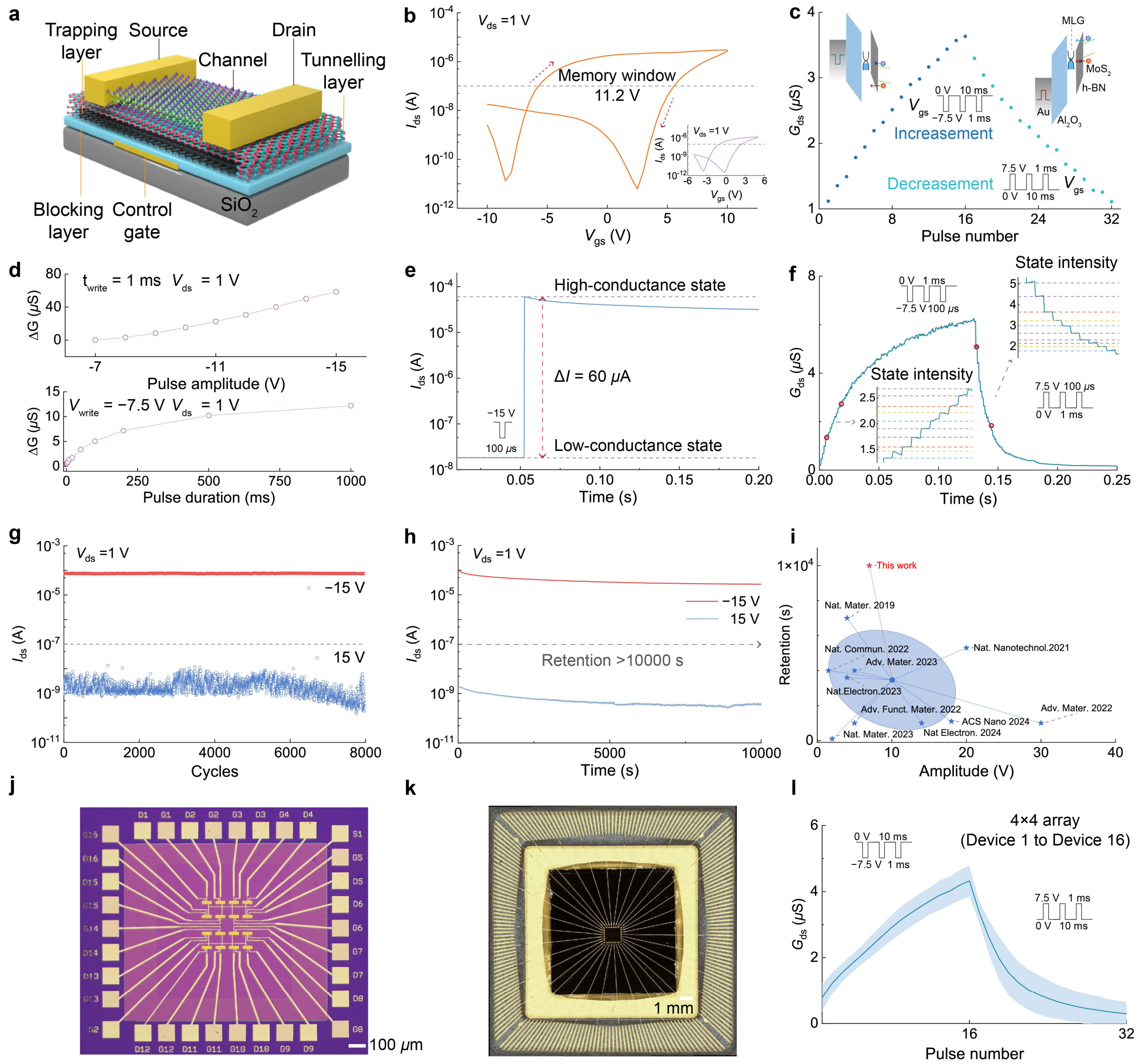

**Figure 2.** Performance characterization of the floating gate synaptic transistor. **a**, Schematic illustration of the floating gate synaptic transistor structure. **b**, Transfer characteristic of the device, with a memory window reaching 11.2 V at $V_{ds}$ = 1 V when $V_{gs}$ is swept from −10 V to 10 V. The inset shows the transfer curve for $V_{gs}$ swept from −5 V to 5 V. **c**, Conductance modulation through 16 negative/positive $V_{gs}$ pulses with the operating mechanism of this synaptic transistor. **d**, Absolute change in conductance (ΔG) as a function of $V_{gs}$ pulse amplitude (top; pulse width = 1 ms) and pulse width (bottom, $V_{gs}$ = −7.5 V). **e**, Switching of conductance state under −15 V $V_{gs}$ pulse of 100 μs. **f**, Conductance changes induced by 100 consecutive negative and positive $V_{gs}$ pulses at 7.5 V amplitude and 100 μs width. The inset depicts multiple conductance states. **g**, Endurance performance of the synaptic transistor executed with alternate positive and negative $V_{gs}$ pulses (±15V, 1 ms). **h**, Retention stability of the synaptic transistor after 10 negative/positive $V_{gs}$ pulse, demonstrating stable high- and low-conductance states. **i**, Comparison with previously reported studies, where the range of the blue ellipse is derived from the mean and standard deviation of the data points in the most advanced research progress in recent years. The centroid represents the mean of the data, while the boundary of the ellipse corresponds to the 95%

confidence interval. Further details can be found in Supplementary Table 1. **j-k**, Optical images of the floating gate synaptic transistor array (j; scale bar, 100 $\mu$m) and its bonded chip (k; scale bar, 1 mm). **l**, Variations in pulse modulation among devices in the 4×4 array. Each device is modulated by 16 consecutive negative/positive $V_{gs}$ pulses with the duration of 1 ms and amplitude of 7.5 V. The blue solid line and blue envelope represent the mean and variance of conductance of 16 devices under $V_{gs}$ pulses.

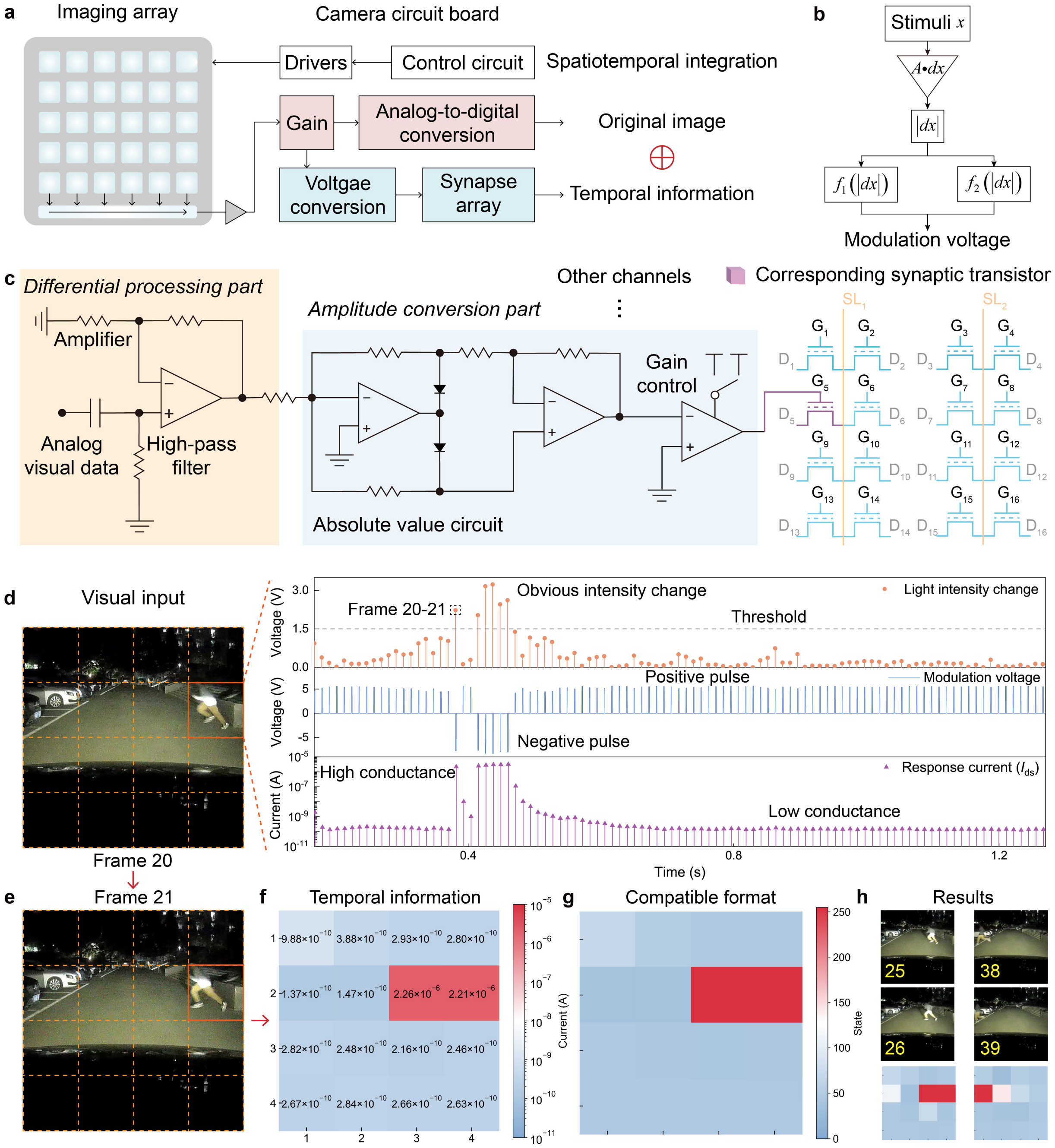

**Figure 3.** **a**, Architecture of the imaging system integrating temporal information with the original image. **b**, Mathematical representation of the circuit used to extract temporal information. **c**, Circuit diagram. **d**, Visual input containing analog visual voltage changes and the corresponding $V_{gs}$ applied to the synaptic transistor at location (2, 4). **e**, Visual input for the subsequent frame following d. **f**, Temporal information derived from the $I_{ds}$ values of the synaptic transistor array states. **g**. Temporal information scaled to the range of 0 to 255. **h**. Examples of processed visual input in the vehicle scenario between frames 25 and 26, and between frames 38 and 39.

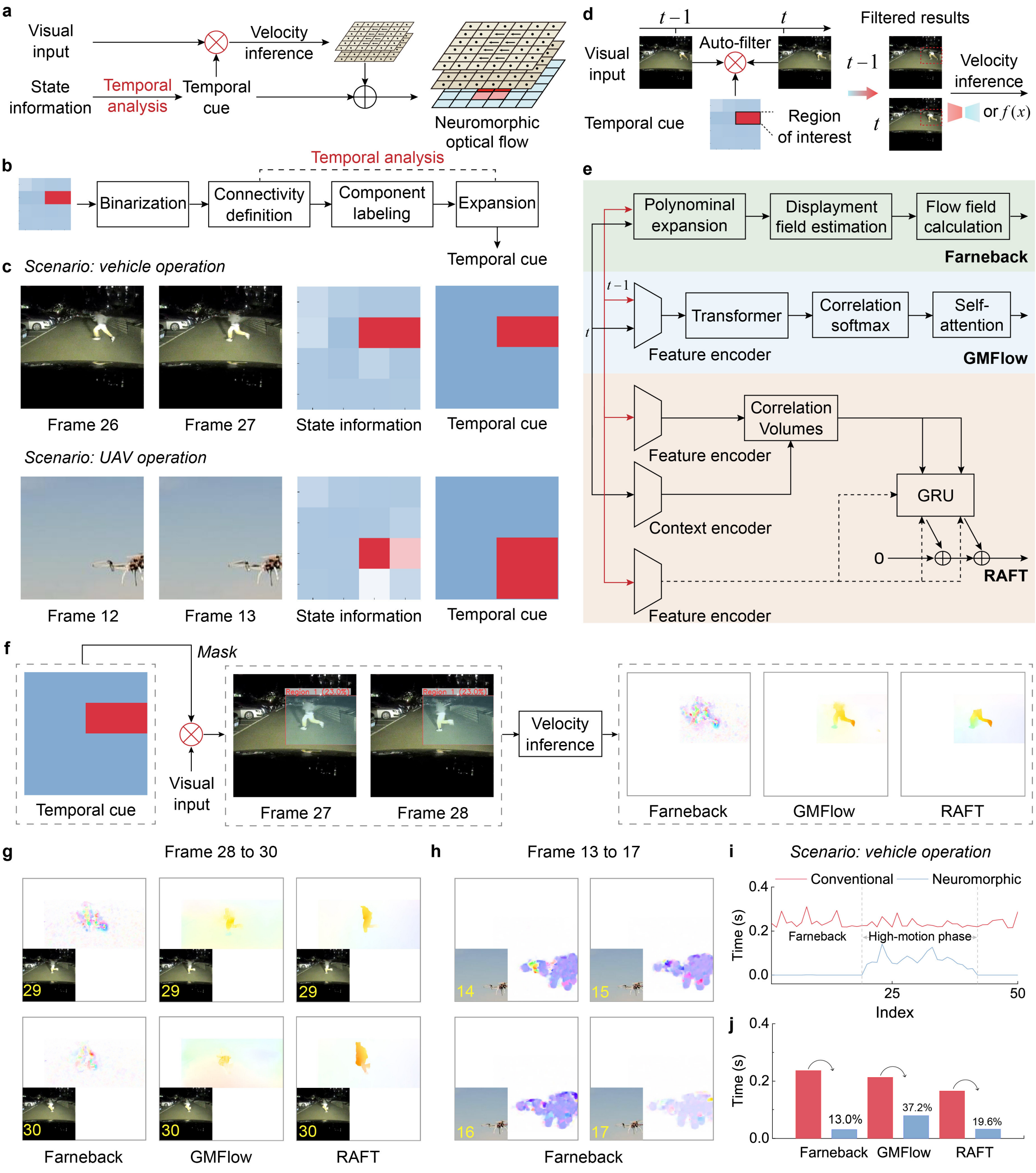

**Figure 4.** **a**, Algorithm structure for generating neuromorphic optical flow. **b**, Process of translating temporal information based on synapse array states into temporal motion cues. **c**, Calculated temporal motion cues for vehicle and UAV operation scenarios. **d**, Accelerated velocity inference using temporal cues. **e**, Schematic diagram of different velocity inference algorithms, including Farneback, GMFlow, and RAFT. **f**, Filtered visual input using temporal cues and resulted movement velocity. **g**, Example results from the vehicle operation scenario. **h**, Example results from the UAV operation scenario. **i**, Comparison of velocity inference times between conventional and neuromorphic optical flow (using Farneback). **j**. Average time comparison.

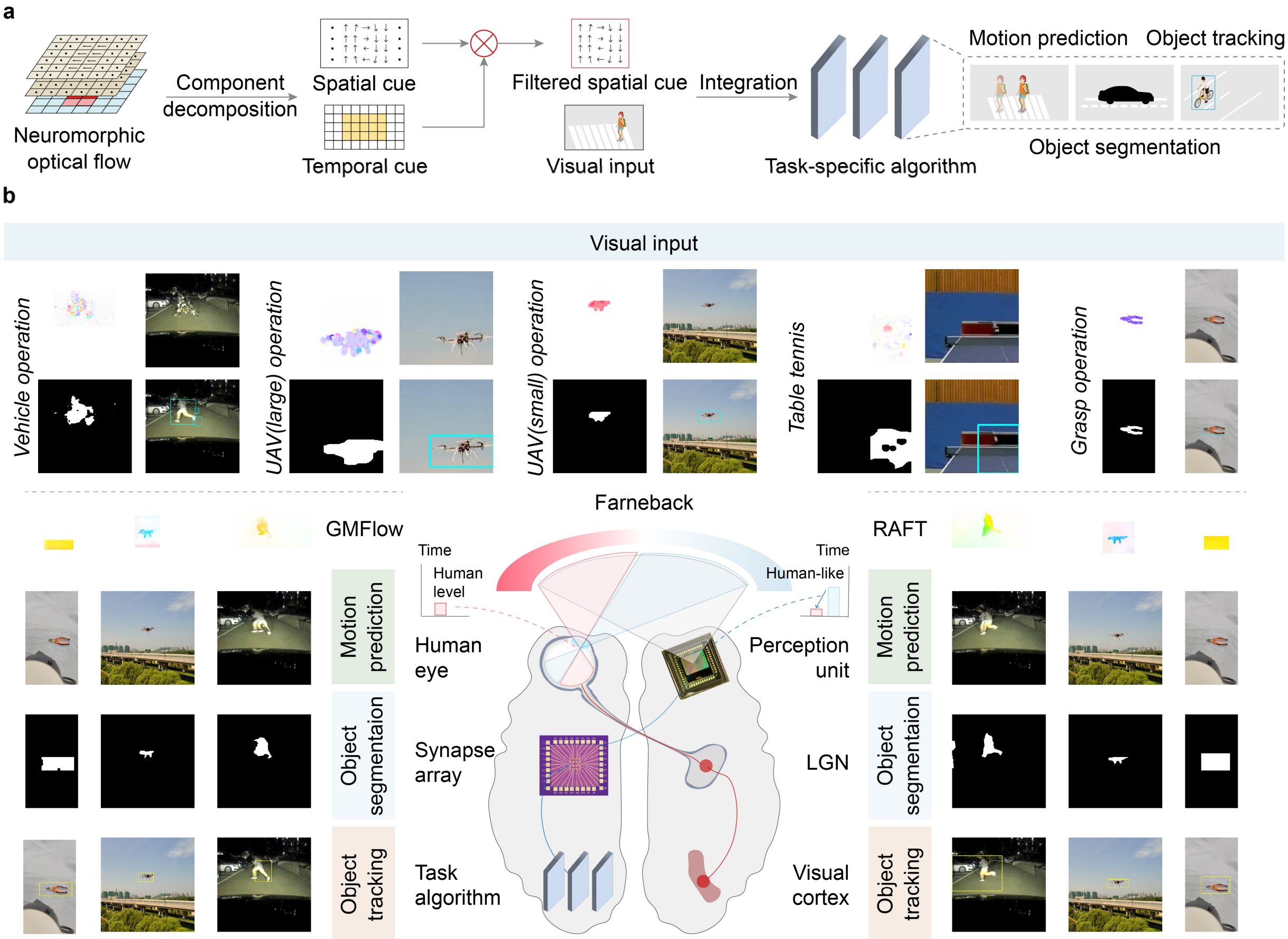

**Figure 5.** **a**, Fundamental visual tasks supported by neuromorphic optical flow. **b**, Processing results of multiple scenarios including vehicle operation, UAV operation, sports activities, and grasp operations.

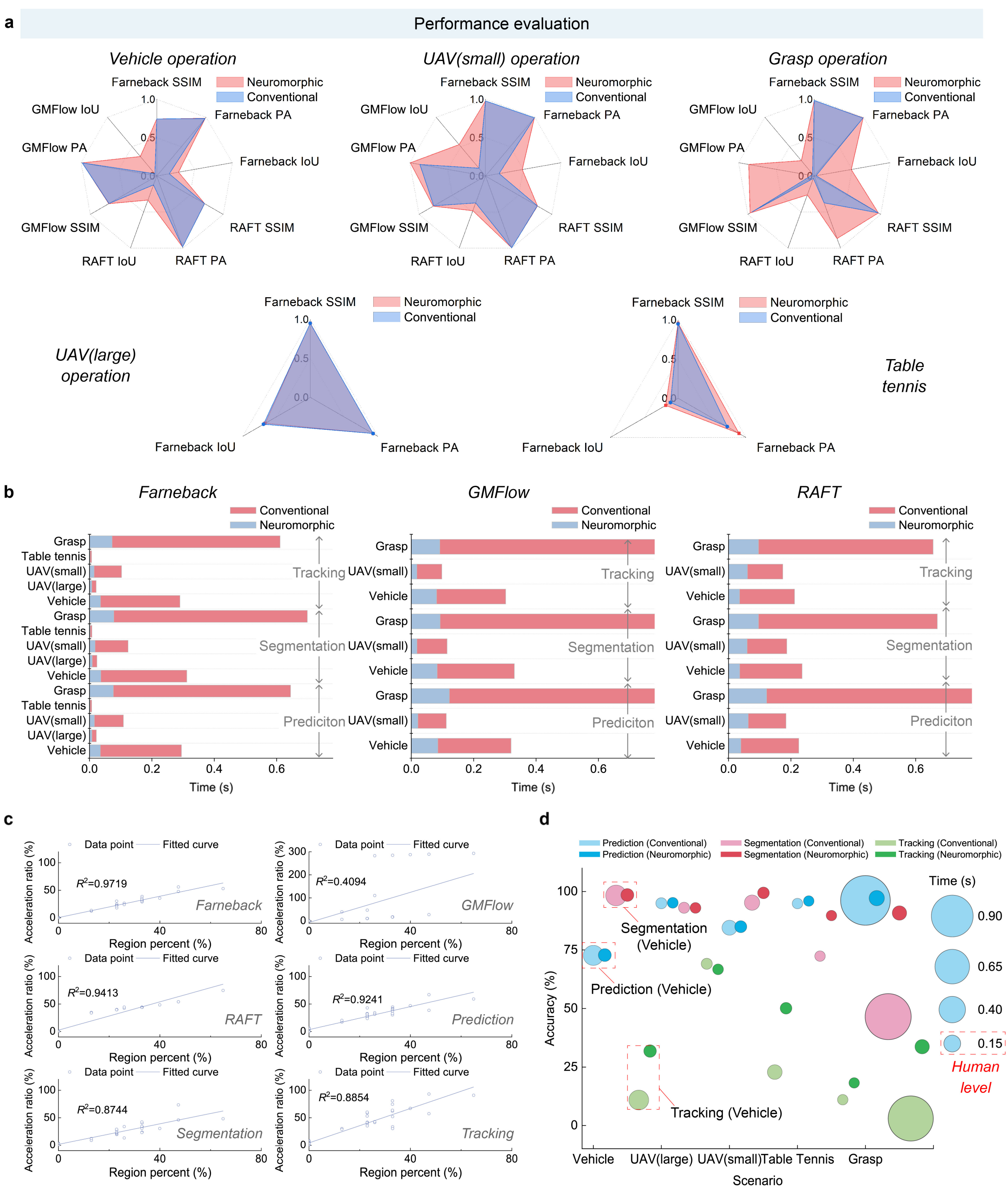

**Figure 6. a**, Task performance comparison between conventional and neuromorphic optical flow. **b**, Comparison of total processing time, including both velocity calculation and task execution, for conventional optical flow versus neuromorphic optical flow. **c**, Acceleration characteristics of our method, illustrating the relationship between acceleration ratio and the percentage of filtered regions based on temporal cues. **d**, Comprehensive comparison of conventional and neuromorphic optical flow across various applications, including average task performance and processing time using different velocity inference methods (Farneback, GMFlow, and RAFT).